\documentclass{article}

\PassOptionsToPackage{numbers}{natbib}
\usepackage[dblblindworkshop, final]{neurips_2025}

\usepackage[utf8]{inputenc} 
\usepackage[T1]{fontenc}    
\usepackage{hyperref}       
\usepackage{url}            
\usepackage{booktabs}       
\usepackage{amsfonts}       
\usepackage{nicefrac}       
\usepackage{microtype}      
\usepackage{xcolor}         

\usepackage{graphicx}
\usepackage{booktabs}

\title{UpstreamQA: A Modular Framework for Explicit Reasoning on Video Question Answering Tasks}
\workshoptitle{Vision Language Models: Challenges of RealWorld Deployment}

\author{
  Jason Nguyen \\
  Lincoln North Star High School \\
  \And
  Ameet Rao \\
  The Charter School of Wilmington \\
  \And
  Alexander Chang \\
  Greenwich High School \\
  \And
  Ishaan Kumar \\
  Santa Susana High School \\
  \AND
  Erin Tan \\
  UC Berkeley
}

\begin{document}

\maketitle

\begin{abstract}
Video Question Answering (VideoQA) demands models that jointly reason over spatial, temporal, and linguistic cues. However, the task’s inherent complexity often requires multi-step reasoning that current large multimodal models (LMMs) perform implicitly, leaving their internal decision process opaque. In contrast, large reasoning models (LRMs) explicitly generate intermediate logical steps that enhance interpretability and can improve multi-hop reasoning accuracy. Yet, these models are not designed for native video understanding, as they typically rely on static frame sampling. We propose UpstreamQA, a modular framework that disentangles and evaluates core video reasoning components through explicit upstream reasoning modules. Specifically, we employ multimodal LRMs to perform object identification and scene context generation before passing enriched reasoning traces to downstream LMMs for VideoQA. We evaluate UpstreamQA on the OpenEQA and NExTQA datasets using two LRMs (o4-mini, Gemini 2.5 Pro) and two LMMs (GPT-4o, Gemini 2.5 Flash). Our results demonstrate that introducing explicit reasoning can significantly boost performance and interpretability of downstream VideoQA, but can also lead to performance degradation when baseline performance is sufficiently high. Overall, UpstreamQA offers a principled framework for combining explicit reasoning and multimodal understanding, advancing both performance and diagnostic transparency in VideoQA in several scenarios.
\end{abstract}

\section{Introduction}

Video Question Answering (VideoQA) is a challenging task that requires large multimodal models (LMMs) to have a comprehensive understanding of video inputs and be capable of answering various questions about them \cite{zhong-etal-2022-video}. These models must be capable of inferring semantic, spatial, temporal, and causal relationships between different entities in the video.

Despite significant advancements, LMMs continue to face substantial limitations regarding weak grounding of question-specific frames, high sensitivity to adversarial perturbations, and disproportionate over-reliance on certain modalities when performing VideoQA tasks \cite{xiao2025videoqaerallmsempirical, mbintvqa}. Traditionally, VideoQA has relied on end-to-end architectures, however their black-boxed nature hinders the transparency of their internal reasoning processes \cite{min2025morevqaexploringmodularreasoning}. 

Recently, a new type of model has been introduced: Large Reasoning Models (LRMs). These models rely on System 2 thinking---logical, deliberate decision-making---as opposed to the quick, intuitive reasoning of System 1 thinking \cite{li202512surveyreasoning}. LRMs can be leveraged in VideoQA to improve accuracy, as they are capable of facilitating intermediate reasoning of questions and incorporating a deeper understanding of spatial, temporal, and causal relationships of video content.

We perform an exploratory study of how explicit reasoning influences VideoQA performance. We introduce a framework for evaluating various upstream tasks processed by LRMs can influence the downstream VideoQA performance. Concretely, our contributions are as follows:
\begin{enumerate}
    \item We introduce UpstreamQA, a novel framework for evaluating explicit reasoning as upstream modules for VideoQA, providing more insight on the intermediate processes in such a complex task.
    \item We perform experiments across two upstream tasks---object identification and scene context generation---as well as two LRMs and two LMMs, reporting results of their effect on VideoQA performance.
    \item We find that although explicit reasoning improves interpretability of logical decision making processes, performance differences are model- and dataset-dependent.
\end{enumerate}

\section{Related Works}
In recent years, Large Multimodal Models (LMMs) that effectively combine visual and linguistic data have driven significant improvements on VideoQA benchmarks \cite{xue2024enhancedmultimodalragllmaccurate}. SOTA approaches on these benchmarks improve factual grounding by incorporating strategies such as Retrieval-Augmented Generation (RAG) \cite{jeong2025videoragretrievalaugmentedgenerationvideo, ren2025videoragretrievalaugmentedgenerationextreme}. However, these systems are commonly trained end-to-end and process the entire task in one continuous pipeline, with no separation between different stages like retrieval, reasoning, and answer generation. This often makes it difficult to analyze why a model fails or succeeds. 

Recently, many approaches have worked to modularize frameworks, including TraveLER \cite{shang2024travelermodularmultilmmagent} and LLoVi \cite{zhang2024simplellmframeworklongrange}, in an attempt to decompose the complex task of VideoQA into several smaller subtasks. These approaches integrate auxiliary large language models (LLMs) to perform tasks such as refining the question or captioning frames \cite{dong2025leadqallmdrivencontextawaretemporal, zhang2024simplellmframeworklongrange, min2025morevqaexploringmodularreasoning}. ENTER is another modular framework that demonstrates better interpretability in the reasoning process through generated event graphs \cite{ayyubi2025entereventbasedinterpretable}. MoReVQA introduces a multi-stage system which produces intermediate outputs applied to specific tasks at each stage \cite{min2025morevqaexploringmodularreasoning}. 

Our framework takes a similar modularized approach to evaluate explicit reasoning on particular upstream tasks. We simplify these multi-stage frameworks into a linear, two-stage process in order to better isolate the effect of the reasoning module on downstream VideoQA performance. Our approach is also designed to be task-agnostic, where the drop-in reasoning module can be evaluated over any upstream reasoning task.

\subsection{Explicit Reasoning in LLMs}
LRMs are a class of LLMs designed to perform multi-step reasoning rather than directly producing answers \cite{li202512surveyreasoning}. These models output intermediate logical reasoning steps to make their thinking process more transparent. Current state-of-the-art (SOTA) LRMs include OpenAI's GPT-5, Google DeepMind's Gemini 2.5 Pro, and Anthropic's Claude Opus 4.1 \cite{openai2025gpt5systemcard, google2025gemini25pushingfrontier, anthropic2025claude_opus_4_1_system_card}.

Recent works demonstrate that enabling explicit reasoning via chain-of-thought, tree-of-thought, or reinforcement learning-based reasoning, substantially boosts task performance and interpretability \cite{xu2025largereasoningmodelssurvey}. This is largely because introducing explicit reasoning facilitates stronger factual grounding and reduces hallucinations. ReasVQA uses a pipeline that generates explicit reasoning, refines the reasoning, then learns from it \cite{liang2025reasvqaadvancingvideoqaimperfect}. They find that noisy or incorrect reasoning hinders performance. Similarly, \citet{10204968} focus on causal reasoning, removing confounders (irrelevant information) by modeling causal graphs. UpstreamQA presents a novel evaluation framework to evaluate explicit reasoning for VideoQA, which, to the best of our knowledge, has not been done.

\section{Methods}
Our method follows a two-stage pipeline. First, we employ reasoning modules to perform distinct upstream video analysis tasks. Specifically, we input 50 uniformly-sampled frames along with an upstream task objective into a multimodal LRM to generate a structured reasoning output. Second, the output is provided to an LMM, that performs the equivalent base VideoQA task but with additional upstream reasoning. We then evaluate the effect of upstream reasoning modules on VideoQA performance using metrics based on accuracy. In this work, we focus specifically on object identification and scene context generation as the upstream tasks, although our framework itself is designed to be task-agnostic. The object identification upstream task focuses on generating a structured inventory of the objects, their attributes (e.g., color, material, etc.), and spatial relationships with one another in a given video \cite{krishna2016visualgenomeconnectinglanguage}. Additionally, the scene context generation upstream task is aimed at recognizing the overall scene category (e.g., kitchen) and generating a comprehensive description of the environment (e.g., environmental details, ambiance, etc.) \cite{NIPS2014_19ea3982, venugopalan-etal-2015-translating}.

\begin{figure}[h]
    \centering
    \includegraphics[width=1\linewidth]{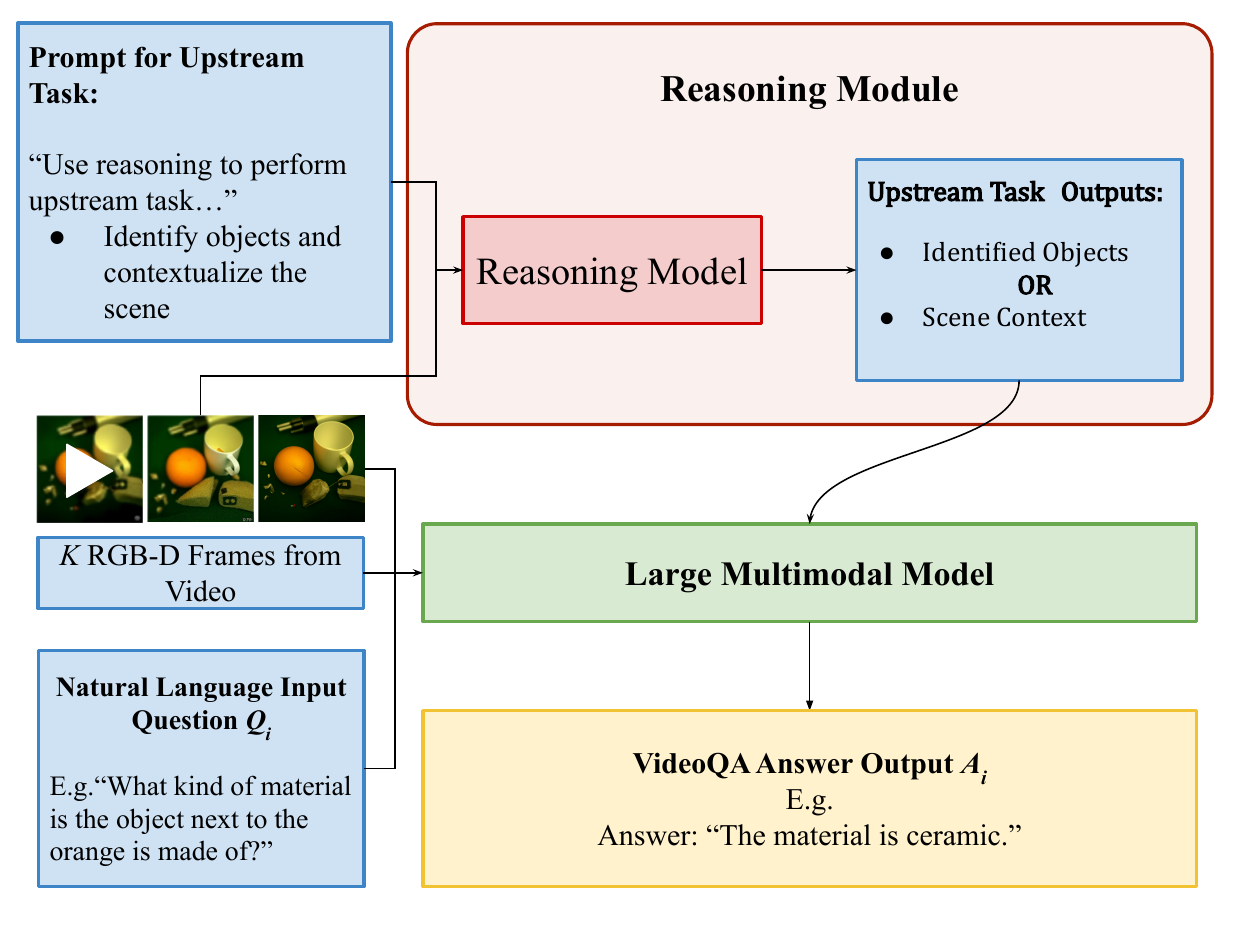}
    \caption{\textbf{Overview of our OpenEQA and NExTQA framework}. An explicit reasoning model is used to perform a specific upstream task, and the output is passed to the LMM along with the original video and question/answers.}
    \label{fig:arch-diagram}
\end{figure}

The outputs of these subtasks are then passed to an LMM along with the original video-question pair, producing our final VideoQA output. Figure \ref{fig:arch-diagram} illustrates the entire pipeline.

\subsection{Datasets}
Our experiments are conducted across two datasets: NExTQA and OpenEQA \cite{xiao2021next, OpenEQA2023}. Both datasets are preprocessed by uniformly sampling 50 frames from each video in the dataset, paired alongside the audio data for the video \cite{openai_cookbook}. The LMM receives a combined input of the sampled frames, audio data, the output from the processed upstream task, and a natural language question prompt. We utilize zero-shot prompting to evaluate the models. All prompts are provided in Section \ref{sec:prompts} of the Appendix.

\paragraph{NExTQA}
NExTQA is a VideoQA dataset containing 5,440 videos and around 52,000 manually annotated question-answer pairs in the form of either multiple-choice QA or open-ended QA. Videos in the dataset show object interactions in daily life, taken from the VidOR dataset \cite{shang2019annotating, thomee2016yfcc100m}.

In this work, we experiment on only the multiple-choice subset, and evaluate performance using accuracy (percentage of correct answers selected). Due to computational constraints, we limited the question set to 2,500 questions, each of which correspond to one of the 298 videos containing the fewest frames in the NExTQA dataset \cite{xiao2021next}. All of the questions from the filtered dataset are between 10-20 seconds long.

\paragraph{OpenEQA} 
OpenEQA is a modern dataset for the task of Embodied Question Answering (EQA), where an agent must navigate a 3D environment and answer natural language questions about it \cite{OpenEQA2023, das2017embodiedquestionanswering}. 

In this work, we specifically use the Episodic Memory EQA (EM-EQA) variant of OpenEQA, which uses episode histories collected from two real-world sources: HM3D and ScanNet \cite{ramakrishnan2021habitatmatterport3ddatasethm3d, dai2017scannetrichlyannotated3dreconstructions}. EM-EQA presents pre-recorded first-person walkthroughs of various environments, which more directly align with a standard VideoQA task, as they consist of continuous sequences that can be processed as standard video input.

In OpenEQA, videos of the given environments are captured in RGB-D frames, allowing for spatial memory. Included are 1,636 human-generated questions regarding 187 scenes based on real-world environments.

We utilize the same evaluation method and correctness metric (LLM-Match) introduced by OpenEQA \cite{OpenEQA2023}. To evaluate outputs, an independent LLM (GPT-4) is used to score outputted answers compared to the ground truth \cite{openai2024gpt4technicalreporta}. Scores are aggregated using the following formula: 
\begin{equation}
C = \frac{1}{N} \sum_{i=1}^{N} \frac{\sigma_i - 1}{4} \times 100\%
\end{equation}
Given a question $Q_i$, a human-annotated answer $A_i^*$, and model output $A_i$, the LLM is prompted to provide a score 
$\sigma_i \in \{1, \dots, 5\}$. 
On this scale, a score of 1 denotes an incorrect response, while a score of 5 denotes a correct response. 
Any intermediate value represents similarity to the upper or lower bound of the scale. 
This process continues for $N$ questions, where $N$ represents the total number of questions.

\subsection{Experiments}
Our experiments are conducted across two different LRMs, two LMMs, and two distinct VideoQA datasets. All models are used off-the-shelf with no additional training or fine-tuning.

\paragraph{Baselines} 
We evaluated baselines by using standalone LMMs. The models used were GPT-4o and Gemini 2.5 Flash \cite{openai2024gpt4ocardb, google2025gemini25pushingfrontier}. Both models take text, images, and audio inputs, while Gemini 2.5 Flash additionally supports video input natively. The baselines are evaluated on ``vanilla" VideoQA, meaning no upstream task outputs are provided. All prompts are provided in section \ref{sec:prompts} of the Appendix.

\paragraph{Experiments} 
We evaluated the addition of explicit reasoning modules by introducing them along with the base LMMs to help with upstream tasks. The reasoning modules used were o4-mini---a smaller, lightweight LRM---and Gemini 2.5 Pro, which is a larger, more powerful LRM \cite{openai2025introducing, google2025gemini25pushingfrontier}. LRMs were evaluated on different upstream tasks and their contribution to performance on downstream VideoQA tasks, being used as a ``drop in" module to assist the LMM. Essentially, additional context is passed into the baseline, providing more information to answer the questions effectively.

\section{Results}

\begin{table*}
\centering
\begin{tabular}{llccc}
\toprule
\textbf{LMM} \rule{0pt}{2ex} & \textbf{LRM} & \textbf{OpenEQA} & \textbf{NExTQA} \\
\midrule
GPT-4o \rule{0pt}{2ex} & ----------- & 67.7 & 62.32\% \\
Gemini 2.5 Flash & ----------- & 58.8 & 78.32\% \\
\\

\textbf{Object Identification} \\
GPT-4o & o4-mini & 55.7 &\textbf{67.48\%} \\ 
GPT-4o & Gemini 2.5 Pro & 59.7 & \textbf{67.08\%} \\
Gemini 2.5 Flash & o4-mini & \textbf{63.6} & 77.44\% \\
Gemini 2.5 Flash & Gemini 2.5 Pro & \textbf{67.1} & 78.00\% \\
\\

\textbf{Scene Context} \\
GPT-4o & o4-mini & 48.1 & \textbf{67.68\%} \\ 
GPT-4o & Gemini 2.5 Pro & 47.8 & \textbf{64.96\%} \\
Gemini 2.5 Flash & o4-mini & \textbf{66.7} & 77.20\% \\
Gemini 2.5 Flash & Gemini 2.5 Pro & \textbf{67.8} & 77.16\% \\
\bottomrule
\end{tabular}
\caption{\textbf{Results on the OpenEQA and NExTQA datasets with distinct LMM and LRM pairs}. The LRMs perform an upstream task of either object recognition or scene context. Performance on OpenEQA is presented through an LLM-Match score, and performance on NExTQA is evaluated by percentage of correct answers. Scores for OpenEQA and NExTQA are provided in the table, with notable score improvements shown in bold.}
\label{tab:mainresults}
\end{table*}

Our results show that the effect of upstream reasoning on downstream VideoQA is dependent on the dataset and base model used. Table \ref{tab:mainresults} reports the overall results from our experiments across all LRM and base model combinations, as well as both upstream tasks. 

On OpenEQA, we find that introducing explicit reasoning significantly improves performance when Gemini 2.5 Flash is used as the base LMM. Specifically, Gemini 2.5 Flash achieves a baseline overall LLM-Match Score of 58.8, which increases to 67.1 and 67.8 when using Gemini 2.5 Flash for object identification and scene context generation, respectively. We observe comparable improvements when using o4-mini as the LRM.

Interestingly, GPT-4o---which has an estimated 1.8 trillion parameters---scored significantly worse than the baseline when combined with an LRM \cite{Lin2025Evaluating}. GPT-4o already exhibited strong performance on standard VideoQA, outscoring Gemini 2.5 Flash by 8.9 points on the baseline.

In NExTQA, we observe performance gains from upstream reasoning in GPT-4o, but not for Gemini 2.5 Flash. GPT-4o achieves a baseline accuracy of 62.32\%, and combining GPT-4o with o4-mini for the object identification upstream task increases the accuracy to 67.48\%. Using Gemini 2.5 Pro as the reasoning model for the same object identification upstream task increases accuracy to 67.08\%. For scene context generation, GPT-4o and o4-mini combined achieve an improved accuracy of 67.68\% (+5.36\%).

In contrast, Gemini 2.5 Flash begins with a baseline accuracy of 78.32\%, much higher than GPT-4o. Combining with the reasoning models for object identification leads to slight degradation in overall accuracy, to 77.44\% for o4-mini (-0.88\% accuracy) and 78.00\% for Gemini 2.5 Pro (-0.32\%). 

These results over both datasets show that introducing upstream explicit reasoning does not always improve downstream VideoQA performance, and that the impact of reasoning is primarily dependent on the dataset and choice of base LMM. When baseline task performance is sufficiently high, we find that introducing explicit reasoning is not always helpful and can even lead to performance degradation.

\subsection{Impacts by Question Type}
In this section, we present additional analysis on OpenEQA to evaluate the performance of the LRMs on their assigned upstream task \cite{OpenEQA2023}. In particular, we hypothesize that the performance of the LRM on the upstream task plays a critical role in determining downstream impacts.

We utilize the change in performance across two question categories---object recognition and world knowledge---as a proxy for the relative performance difference between the base LMM and the LRM on the relevant upstream task. OpenEQA does not provide ground-truth labels for object identification or scene context generation; instead, questions are categorized into seven distinct question types which can be used to evaluate a model's performance on specific facets of VideoQA. Object recognition questions evaluate a model's ability to recognize objects that appear in the episodic history, whereas world knowledge questions assess a model's ability to leverage external background knowledge about the world \cite{OpenEQA2023}. We select object recognition and world knowledge because they most directly align with our upstream tasks of object identification and scene context generation, respectively. In our analysis, we stratify LLM-Match scores for the object recognition and world knowledge question types to better isolate the effects that reasoning modules have on OpenEQA performance.

\begin{figure}[h]
    \centering
    \includegraphics[width=1\linewidth]{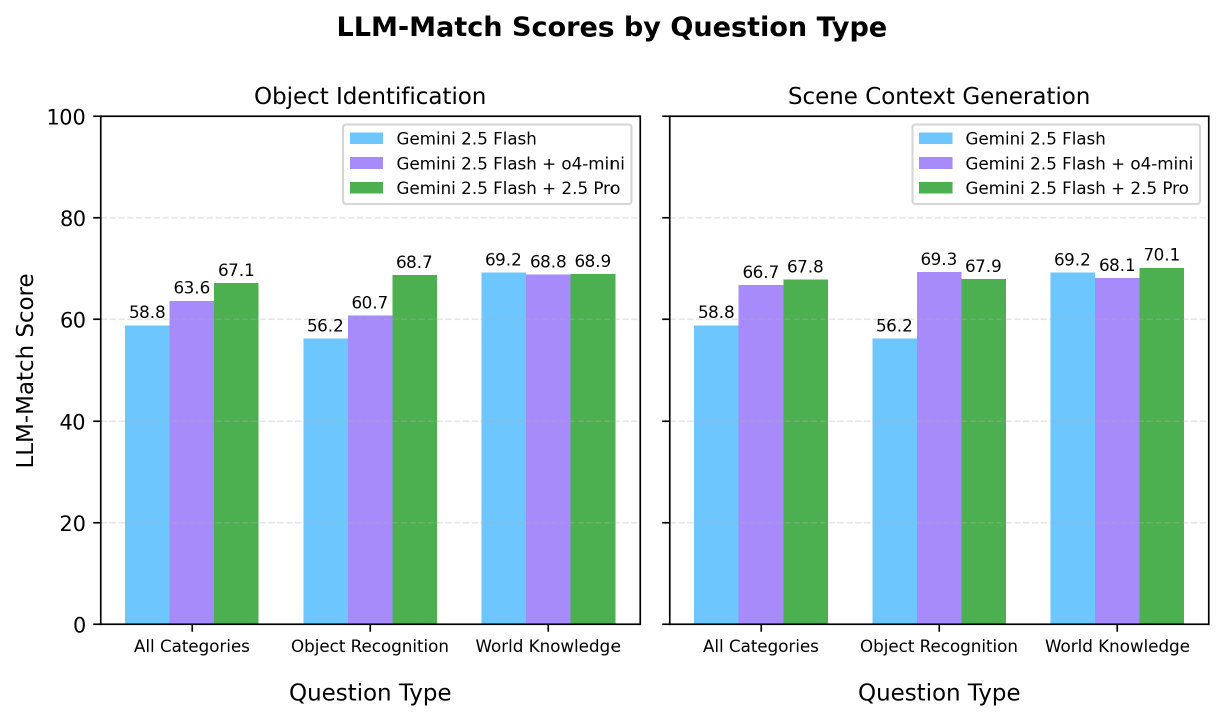}
    \caption{\textbf{LLM-Match Scores stratified by question category on OpenEQA using Gemini 2.5 Flash as the base model}. Results are reported for both upstream tasks: object identification (left) and scene context generation (right). Upstream reasoning improves overall accuracy and accuracy on object recognition questions while world knowledge question performance remains relatively stagnant.}
    \label{fig:openeqa-bar-chart}
\end{figure}

Figure \ref{fig:openeqa-bar-chart} reports the results stratified by question type for Gemini 2.5 Flash on OpenEQA. The full set of results for both models are presented in Table \ref{tab:extended-results-table} in the Appendix. For object identification, we find that performance on object recognition questions improves significantly when reasoning is used. As a control, we also provide the change in scores for world knowledge questions when object identification is the upstream task and find that performance stays relatively stagnant. Curiously, we observe similar trends for scene context generation, wherein object recognition performance improves while world knowledge remains stagnant. These results suggest that the benefit of explicit upstream reasoning largely lies in its ability to provide factual grounding and not so much \textit{understanding}; for more structured questions such as object recognition, reasoning is helpful, whereas it may not be as helpful for broader questions about world knowledge.

\section{Discussion}
In this work, we introduce a novel two-stage framework for evaluating explicit reasoning in VideoQA across the NExTQA and OpenEQA datasets. The modularity of our framework allows for greater flexibility, allowing for experimentation with different model and task combinations. Furthermore, our framework also allows for greater interpretability of the models and by exposing intermediate reasoning steps rather than relying solely on end-to-end evaluation metrics.

The results of our experiments reveal the effect of our framework on VideoQA accuracy on certain tasks, while leading to performance degradation on others. This indicates that selective integration of modular reasoning via UpstreamQA can yield substantial improvements in task-specific VideoQA performance.

\paragraph{Limitations \& Future Work}
Our work presents promising preliminary findings for better understanding the role of explicit reasoning models in improving complex tasks like VideoQA. Additional experiments are necessary to account for various confounding factors, such as upstream task performance. The scope of our evaluation was limited to zero-shot VideoQA, although comparisons to other methods such as chain-of-thought prompting are also necessary.

Future expansions of this work may consider exploring further usage of modular reasoning to encompass other core video reasoning components and their effect on VideoQA performance. Moreover, subsequent research may additionally investigate why performance degradation occurs on certain models while significant performance improvements are observed on others.

\bibliographystyle{unsrtnat}
\bibliography{neurips_2025}

\appendix

\section{Extended Results} \label{sec:extended-results}

\begin{table}[h]
\centering
\begin{tabular}{llccc}
\toprule
\textbf{LMM} \rule{0pt}{2ex} & \textbf{LRM} & \textbf{Overall} & \textbf{Object Recognition} & \textbf{World Knowledge} \\
\midrule
GPT-4o \rule{0pt}{2ex} & ----------- & 67.7 & 71.4 & 62.4 \\
Gemini 2.5 Flash & ----------- & 58.8 & 56.2 & 69.2 \\
\\

\textbf{Object Identification} \\
GPT-4o & o4-mini & 55.7 & 56.7 & 60.7 \\ 
GPT-4o & Gemini 2.5 Pro & 59.7 & 62.6 & 64.9 \\
Gemini 2.5 Flash & o4-mini & \textbf{63.6} & \textbf{60.7} & 68.8 \\
Gemini 2.5 Flash & Gemini 2.5 Pro & \textbf{67.1} & \textbf{68.7} & 68.9 \\
\\

\textbf{Scene Context} \\
GPT-4o & o4-mini & 48.1 & 43.7 & 54.8 \\ 
GPT-4o & Gemini 2.5 Pro & 47.8 & 43.1 & 55.5 \\
Gemini 2.5 Flash & o4-mini & \textbf{66.7} & \textbf{69.3} & 68.1 \\
Gemini 2.5 Flash & Gemini 2.5 Pro & \textbf{67.8} & \textbf{67.9} & 70.1 \\
\bottomrule
\addlinespace[.5em]
\end{tabular}
\caption{\textbf{LLM-Match scores on the OpenEQA dataset with distinct LMM and LRM pairs}. The LRMs perform an upstream task of either object recognition or scene context. The scores are for all questions (overall), object recognition questions, and world knowledge questions. The object recognition and world knowledge questions are stratified from the OpenEQA dataset, which originally has seven questions types. Both question types are stratified since we hypothesize that they will display the most significant score differences through our framework. Notable score improvements are represented in bold.}
\label{tab:extended-results-table}
\end{table}

\section{Prompts} \label{sec:prompts}
In this section we present the prompts used in our experiments. Figure \ref{fig:base-prompt} represents the baseline prompt passed into GPT-4o and Gemini 2.5 Flash without any upstream task augmentation. \{question\} in Figure \ref{fig:base-prompt} denotes where the question sourced from the dataset is input. 

The prompts for the two upstream tasks, object identification and scene context generation, are presented in Figures \ref{fig:object-identification-prompt} and \ref{fig:scene-context-prompt}, respectively. Figure \ref{fig:upstream-qa} shows the prompt used for processing QA questions along with upstream reasoning, which is passed into the LMM. In figure \ref{fig:upstream-qa}, \{upstream\_task\} represents the upstream input, \{question\} represents the question sourced from the dataset, and \{upstream\_task\_placeholder\} represents a task-differentiating variable, since the same prompt was used for different upstream tasks.

\begin{figure}[h]
    \centering
    \includegraphics[width=1\linewidth]{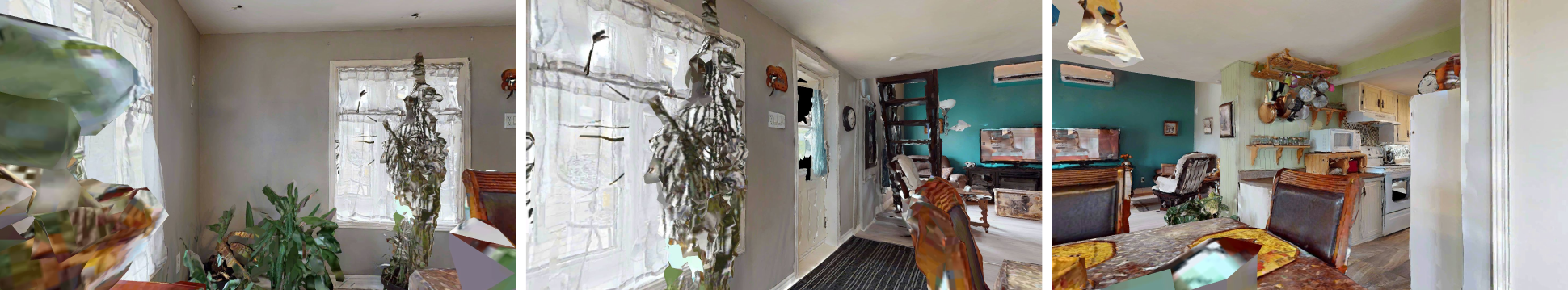}
    \fbox{%
        \parbox{1\linewidth}{%
            {\fontsize{10pt}{12pt}\selectfont\textbf{Example Scene Context Output}}\\
            {\fontsize{8pt}{10pt}
                Scene Overview:\\
                Identification and Purpose\\
                Open-plan living space combining a small dining area, a kitchenette, and a lounge.\\
                Functions include meal preparation and dining (nearest the kitchen), casual seating and media viewing (farther in), and general circulation to a loft via a narrow staircase.\\[4pt]
                Architectural Details\\
                Flooring: Light-toned, wide-plank wood or wood-look laminate laid continuously throughout.\\
                Walls: Predominantly light gray paint with white trim; one full accent wall in the lounge is painted teal.\\
                Ceiling: Flat, painted white over the dining and kitchen zone; vaulted wood-paneled section over the lounge.\\
                Windows \& Doors:\\
                Two double-hung windows dressed with semi-sheer lace curtains in the dining corner.\\
                A white paneled exterior door (with glazed upper section) opens near a side wall.\\
                Interior doorways (one near the lounge leads to a bedroom) feature white trim.\\
                A shallow closet with bifold white doors sits adjacent to the kitchen entrance.\\
                Staircase: A compact, dark-stained wooden ladder-style stair rises against the teal wall to a loft or mezzanine level.\\
                Kitchen Fixtures: Light oak cabinetry, open shelving for glassware and cookware, a wall-mounted pot rack made of rough-hewn wood, white range and refrigerator, microwave on a countertop.\\[4pt]
                Ambiance\\
                Lighting: Bright natural daylight filters through the lace curtains; ambient light is supplemented by a pendant lamp over the dining table and floor lamps in the lounge.\\
                Furnishings \& Decor:\\
                Dining: A small rectangular table with two high-back chairs and leaf-patterned placemats.\\
                Lounge: Two upholstered glider chairs, a leather-look loveseat, a low TV stand with flat-screen, and a rustic storage trunk used as a coffee table.\\
                Accessories: Several potted plants cluster by the window, framed prints on the walls, a wall clock, and a decorative wall plaque.\\
                Condition \& Style: The space feels cozy and lived-in, leaning toward a rustic-eclectic style. It is generally tidy, though numerous decorative items and plants create a slightly busy, homey atmosphere.
            }
        }
    }
\caption{Example output of generated \textbf{scene context}.}
\label{fig:scene-context-output}
\end{figure}

Example outputs for an object identification or scene context generation task are presented in Figures \ref{fig:object-identification-output} and \ref{fig:scene-context-output}, respectively.


\begin{figure}[h]
    \centering
    \fbox{%
        \parbox{1\linewidth}{%
            {\fontsize{10pt}{12pt}\selectfont\textbf{Baseline VideoQA Prompt}}\\
            {\fontsize{8pt}{10pt}
            You are an embodied AI assistant. Your task is to answer a question about a given environment from images using your own knowledge.\\
            **Primary Goal:** Apply your general and functional knowledge to reason about what is possible or true based on the context using your given inputs.\\
            
            **CRITICAL RULES:**
            - **Be direct and brief.** Your answer should be as short as possible.\\
            - For questions that can be answered with "Yes" or "No", you MUST answer with only "Yes" or "No".\\
            - **Do not explain your reasoning or mention the provided context.** Avoid phrases like "Based on the context..." or "Observing the images...".\\
            - **Don't overanalyze, many of the answers are simple and are not extremely detailed or have a lot of adjectives**\\
            - **Very short statements are also acceptable**\\

            **User Query:** \{question\}
            A:
        }
    }
}
\caption{Prompt for standalone LMMs to run baselines.}
\label{fig:base-prompt}
\end{figure}

\begin{figure}[h]
    \centering
    \fbox{%
        \parbox{1\linewidth}{%
            {\fontsize{10pt}{12pt}\selectfont\textbf{Object Identification Prompt}}\\
            {\fontsize{8pt}{10pt}
            *Note* This is an upstream task of object identification and their spatial layout.\\
            Use reasoning to analyze the provided sequence of images from a first-person perspective. Your goal is to generate a comprehensive, structured, and factual description of the object inventory and spatial layout. Be as detailed as possible.\\
            Provide your analysis in the following structured format:\\[4pt]
            Object Inventory:\\
            Major Items: List prominent furniture and appliances. For each item, specify its attributes (color, material, shape) and its location relative to the room and other objects (e.g., "A rectangular wooden desk is against the far wall, to the left of the window").\\
            Minor Items: Detail smaller objects such as decorations, electronics, containers, or personal items found on surfaces or shelves. Describe their key features and placement.
        }
    }
}
\caption{Upstream task prompt for LRMs to \textbf{identify objects} in the inputted frames.}
\label{fig:object-identification-prompt}
\end{figure}

\begin{figure}[h]
    \centering
    \fbox{%
        \parbox{1\linewidth}{%
            {\fontsize{10pt}{12pt}\selectfont\textbf{Scene Context Prompt}}\\
            {\fontsize{8pt}{10pt}
            *Note* This is an upstream task of scene context.\\
            Use reasoning to analyze the provided sequence of images from a first-person perspective. Your goal is to generate a comprehensive, structured, and factual description of the scenes. Be as detailed as possible.\\
            Provide your analysis in the following structured format:\\[4pt]
            Scene Overview:\\
            Identification and Purpose: Identify the type of room or space (e.g., office, bedroom, kitchen). Describe its likely purpose.\\
            Architectural Details: Note the materials and styles of the floor, walls, and ceiling. Mention significant structural elements like windows, doors, or stairs.\\
            Ambiance: Briefly describe the overall condition and atmosphere (e.g., tidy, cluttered, modern, rustic, well-lit, dimly lit).
        }
    }
}
\caption{Upstream task prompt for LRMs to generate \textbf{scene context} of the inputted frames.}
\label{fig:scene-context-prompt}
\end{figure}

\begin{figure}[h]
    \centering
    \fbox{%
        \parbox{1\linewidth}{%
            {\fontsize{10pt}{12pt}\selectfont\textbf{VideoQA Prompt with Included Upstream Reasoning}}\\
            {\fontsize{8pt}{10pt}
            You are an embodied AI assistant. Your task is to answer a question by synthesizing information from images, an upstream task, and your own knowledge.\\

            **Primary Goal:** Use the upstream task as a helpful source for the current state of the environment.\\
            **Secondary Goal:** Apply your general and functional knowledge to reason about what is possible or true based on the context using your given inputs.\\
            
            **CRITICAL RULES:**\\
            - Be direct and brief. Your answer should be as short as possible\\
            - For questions that can be answered with "Yes" or "No", you MUST answer with only "Yes" or "No"\\
            - Do not explain your reasoning or mention the provided context. Avoid phrases like "Based on the context..." or "Observing the images..."\\
            - Don't overanalyze, many of the answers are simple and are not extremely detailed or have a lot of adjectives\\
            - Very short statements are also acceptable\\
            - The upstream task is designed to help you answer questions—not give you the answer\\
            - Important information may be missing from the upstream task, so make sure you still use your general and functional knowledge to reason\\
            
            **Upstream Task:**\\
            This is the upstream task of \{upstream\_task\_placeholder\} for the given scenes.\\
            \{upstream\_task\}\\
            
            **User Query:** \{question\}\\
            A:
        }
    }
}
\caption{Upstream task prompt for LRMs to generate \textbf{scene context} of the inputted frames.}
\label{fig:upstream-qa}
\end{figure}

\begin{figure}[h]
    \centering
    \fbox{%
        \parbox{1\linewidth}{%
            {\fontsize{10pt}{12pt}\selectfont\textbf{Example Object Identification Output}}\\
            Based on the sequence of images, here is a detailed analysis of the object inventory and spatial layout of the interconnected kitchen, dining, and living areas. \\
            Object Inventory \\\\ Major Items: \\ Refrigerator: A tall, white refrigerator is located at the edge of the kitchen area, to the left of a bifold closet door.\\   Stove/Oven: A white, freestanding electric range with an oven is positioned to the right of the main kitchen counter space.\\
            Dining Table: A large rectangular dining table with a mottled brown and gray tabletop sits between the kitchen and living areas.\\ Dining Chairs: At least two wooden dining chairs with dark brown frames, brown leather-like upholstered seats, and high, slatted backs are positioned at the dining table.\\ Sofa: A plush, brown fabric sofa with multiple throw pillows is situated against the far right wall of the living room.\\ Armchair: A matching plush, brown fabric armchair is located in the corner to the left of the TV console.\\ Rocking Chairs: Two rocking chairs with dark wood frames are present. One, with a white/beige cushioned seat and a white throw blanket, is next to the sofa. Another, with similar light-colored cushions, is next to a tall shelving unit.\\ TV Console / Fireplace: A dark brown wooden media console with glass-front cabinets and a central electric fireplace insert is placed against the teal accent wall, beneath the television.\\ Tall Shelving Unit: A very tall, dark wood, open-back shelving unit with multiple shelves stands against the gray wall, acting as a partial divider between the dining and living areas.\\ Trunk Coffee Table: A vintage-style rectangular trunk, light-colored with a dark printed design (possibly a map or historical illustration), serves as a coffee table in front of the TV console.\\ Kitchen Cabinets: Off-white or cream-colored kitchen cabinets are visible, with a set of upper and lower cabinets to the right of the stove.\\\\ Minor Items: \\ Microwave: A white microwave oven sits on a small wooden stand on the kitchen countertop.\\ Pot Rack: A large, wooden, ceiling-mounted pot rack is installed above the kitchen peninsula. Numerous pots and pans, including copper and stainless steel ones, hang from it.\\ Wall Shelves:** Two wooden shelves are mounted on the light green kitchen wall below the pot rack. The lower shelf holds a neat row of clear glass jars with lids.\\ Potted Plants: Several potted plants are throughout the space. A large, leafy green plant is in the corner of the dining area by a window. Another tall, thinner plant is next to it. A smaller plant is on the floor near the kitchen peninsula.\\ Television: A large, widescreen flat-panel TV is mounted on the teal accent wall above the media console.\\ Air Conditioner: A white, ductless mini-split air conditioner unit is mounted high on the teal wall, above and to the right of the TV.\\ Lighting: A floor lamp with a silver-toned base and a white upward-facing shade stands between the tall shelving unit and the armchair. A ceiling-mounted light fixture with a bell-shaped glass shade is visible above the dining area.\\ Clocks: A round, black-rimmed analog clock is on the gray wall near the exterior door. Another round, white-faced clock is visible on top of the refrigerator.\\ Wall Decor: Several framed pictures adorn the walls. Two small, square-framed pictures are on the teal wall. A larger, rectangular framed picture of a winter scene hangs on the off-white wall next to a doorway. A small, irregularly shaped piece of wood art is on the gray wall by the dining area.\\   Curtains: Sheer, white, gathered curtains are on the windows in the dining area. Darker gray or black curtains are on the window in the living area corner.\\ Small Tables: A small, slatted wooden coffee table is next to the armchair by the tall shelf. A small end table with a crisscross base is next to the sofa.\\ Rugs: A small, dark, striped rug is on the floor by the exterior door. A small area rug with a dark border is under the trunk coffee table.\\ Closet Door: A white, two-panel bifold door is located between the refrigerator and the dining area wall.,
            {\fontsize{8pt}{10pt}

            }
        }
    }
\caption{Example output of generated \textbf{object identification}.}
\label{fig:object-identification-output}
\end{figure}

\end{document}